\pdfoutput=1

\documentclass[11pt]{article}

\usepackage[]{EMNLP2023}

\usepackage{times}
\usepackage{latexsym}
\usepackage{graphicx,multirow}

\usepackage{url}

\usepackage{subfigure}

\usepackage{xcolor}

\definecolor{purple}{rgb}{0.5,0,1}
\definecolor{dcyan}{rgb}{0.2,0.6,0.5}
\definecolor{light-gray}{gray}{0.95} 
\definecolor{darkgreen}{RGB}{0,140,0}
\definecolor{darkred}{RGB}{200,0,0}
\definecolor{lightgreen}{RGB}{189,252,192}
\definecolor{lightred}{RGB}{255,205,212}
\definecolor{lightyellow}{RGB}{255,240,160}
\definecolor{lightblue}{RGB}{195,221,255}
\definecolor{lightpurple}{RGB}{232,209,255}

\usepackage[T1]{fontenc}

\usepackage[utf8]{inputenc}

\usepackage{microtype}

\usepackage{inconsolata}

\title{Exploring the Numerical Reasoning Capabilities of Language Models: \\A Comprehensive Analysis on Tabular Data}

\author{Mubashara Akhtar\textsuperscript{1}\thanks{~~Equal contributions}~~, Abhilash Shankarampeta\textsuperscript{2*}, Vivek Gupta\textsuperscript{3}, Arpit Patil\textsuperscript{4} \\ {\bf Oana Cocarascu\textsuperscript{1}} and {\bf Elena Simperl\textsuperscript{1}} \\[5pt]
        \textsuperscript{1}King's College London
        \textsuperscript{2}Meesho
        \textsuperscript{3}University of Pennsylvania
        \textsuperscript{4}University of Utah\\[5pt]
        \texttt{mubashara.akhtar@kcl.ac.uk}\\
        \texttt{abhilash.shankarampeta@meesho.com}} 

\begin{document}
\maketitle
\begin{abstract}

Numbers are crucial for various real-world domains such as finance, economics, and science. Thus, understanding and reasoning with numbers are essential skills for language models to solve different tasks. While different numerical benchmarks have been introduced in recent years,
they are limited to specific numerical aspects mostly. 
In this paper, we propose a hierarchical taxonomy for numerical reasoning skills with more than ten reasoning types across four levels: representation, number sense, manipulation, and complex reasoning. 
We conduct a comprehensive evaluation of state-of-the-art models 
to identify reasoning challenges specific to them.
Henceforth, we develop a diverse set of numerical probes employing a semi-automated approach. 
We focus on the tabular Natural Language Inference (TNLI) task as a case study and measure models' performance shifts. 
%
Our results show that no model consistently excels across all numerical reasoning types. Among the probed models, FlanT5 (few-/zero-shot) and GPT-3.5 (few-shot) demonstrate strong overall numerical reasoning skills compared to other models. Label-flipping probes indicate that models often exploit dataset artifacts to predict the correct labels.\footnote{Data and code are available at \url{https://github.com/mubasharaak/numerical_reasoning}.}\\

\if 

Vivek Gupta Comments
Dataset Level
1. Counterfactual Tables
2. More Datasets Addition (Recasting)
3. Tabular QA tasks
4. Reasoning on Complex and Manipulation

Model Level
1. More Models specially Numerical and LLM based (GPT-4 also)
2. [Imp] We need to do some details analysis on consistency and confusion aspect. We need to also think of better analysis approches/metrics we could work around.

Writing Level
1. Structure of the paper need improvement
2. Content writing need to improve

Talk to me in a one hour call I will tell details
Add anything which I missed in this.

\fi

\end{abstract}

\section{Introduction}
\label{sec:introduction}


Numerical data is ubiquitous in the real-world. Many applications in domains such as finance, economics and science require understanding and reasoning with numbers. 
In recent years, benchmarks were introduced to study language models' numeracy skills~\citep{zhang-etal-2020-language-embeddings, wallace-etal-2019-nlp, dua-etal-2019-drop}. 
However, these datasets
mostly concentrate on few, specific numerical reasoning types (e.g. scales~\citep{zhang-etal-2020-language-embeddings}). 

\begin{table}[t]
\begin{center}
\small{
\begin{tabular}{l l}\\
\hline
\multicolumn{2}{c}{\textbf{Hulk}} \\
\hline
\textbf{Directed by} & Ang Lee \\
\textbf{Release date} & June 20, 2003 \\
\textbf{Running time} & 138 minutes\\
\textbf{Budget} & $\$137$ million \\
\textbf{Box office} & $\$245.4$ million\\
\hline
\end{tabular}}
\end{center}
\begin{tabular}{ c c }
    \fbox{\begin{minipage}{18.8em}
            \small       
              \textbf{H$1$:} Hulk was released on $20$th June, $2003$. \textit{(E)}\\
              \textbf{Date:} Hulk was released on 20-06-2003. \textit{(E)}\\
              \textbf{Date Flip:} Hulk was released on 12-08-2009. \textit{(C)}\\
           \rule{\linewidth}{0.01em}
              \textbf{H$2$:} The movie has a length of $138$ minutes. \textit{(E)}\\
              \textbf{Appr:} The movie has a length of about $150$ minutes. \textit{(C)}\\\
           \rule{\linewidth}{0.01em}
              \textbf{H$3$:} The movie can be watched in about two hours. \textit{(E)}\\\
              \textbf{Num:} The movie can be watched in about $2$ hours. \textit{(E)}\\\
              \textbf{Num Flip:} The movie can be watched in $1$ hours. \textit{(C)}\\
           \rule{\linewidth}{0.01em}
              \textbf{Arith:} Hulk brought in $\$108.4$ million profit. \textit{(E)}\\
              \textbf{Arith Flip:} Hulk brought in $\$120.9$ million profit. \textit{(C)}
    \end{minipage}} & 
\end{tabular}
\vspace{-0.5em}
\caption{\small \label{table:intro} Base hypotheses (H$1$, H$2$, H$3$) and (\textbf{flip}ped) probes for heterogeneous numbers (i.e. \textbf{date}), \textbf{appr}oximation, \textbf{num}eracy, and \textbf{arith}metic. Labelled as \textbf{E}ntail or \textbf{C}ontradict.} 
\vspace{-1.0em}
\end{table}

Moreover, evaluating models on numerical benchmarks, it often remains unclear why models struggle with the tasks. For example, the issues can arise from models struggling to recognize numerical representations in text, failing to compute arithmetic operations, or predicting incorrect outputs due to a lack of numerical commonsense knowledge. We aim to explore these questions in greater detail in this study.
Limitations of language models' numerical abilities, as discussed in prior research, include tokenization and representation of numbers in text~\citep{thawani-etal-2021-representing}, 
hallucination~\citep{ji-etal-2023-hallucination, chen-etal-2023-purr, ye-etal-2023-decomposers}, and generalizability/robustness  issues~\citep{razeghi-etal-2022-impact, geva-etal-2020-injecting, xu-etal-2022-towards-robust}.

Successful numerical reasoning requires a combination of skillsets: understanding representation of numbers~\citep{thawani-etal-2021-numeracy,thawani-etal-2021-representing} and their meaning in a given context~\citep{loukas-etal-2022-finer}, applying operations~\citep{geva-etal-2020-injecting,patel-etal-2021-nlp}, and integrating factual and commonsense numerical knowledge to solve numerical problems~\citep{lin-etal-2020-birds,park-etal-2022-language}. 
For example, classifying the hypotheses \textit{``The movie can be watched in about $2$ (or `two') hours.''} from Table~\ref{table:intro} requires understanding that both \textit{``$2$''} and \textit{``two''} depict the same numerical value, converting \textit{``$2$ hours''} to another unit (i.e. $120$ minutes), and applying approximation to map \textit{``$120$ minutes''} to \textit{``$138$ minutes''} 
in the table. 


In this paper, we evaluate state-of-the-art models with diverse architectures, sizes, and training settings.
To assess which reasoning types are challenging for specific models, we create a diverse and large set of numerical probes and measure shifts in models' performance. 
We organize all probed reasoning types in 
a hierarchical taxonomy (Figure~\ref{fig:numreasoning}).
Inspired by how humans understand and reason with numbers, as well as previous numerical benchmarks, we include eleven reasoning types across four level: \emph{representation}, \emph{number sense}, \emph{manipulation}, and \emph{complex reasoning}. 
We apply a semi-automated approaches for probe creation.
We select tabular NLI (TNLI) as a case study task, given three criteria: $(i)$ numerical data (numbers, percentages, dates, etc.) is prevalent in tables; $(ii)$ tables are common in real-world data sources such as in scientific publications, database systems and financial documents; $(iii)$ tables as structured data facilitate automated perturbations to create large-scale probing sets. 
See Table~\ref{table:intro} for some examples of probes created from hypotheses (H$1$, H$2$, H$3$) and the given table. 

Our experiments conclude that large language models (LLMs) like FlanT$5$ and GPT$3.5$ perform better than other models on various numerical reasoning tasks. 
Both table-based and numerical models struggled to understand data with flipped labels and negative values.
Moreover, we observe that some models' performance improves significantly for counterfactual probes (e.g. NT$5$ and TAPAS) and label-flipping probes (e.g. FlanT$5$ zero-shot), which indicates that models might exploit dataset artifacts and are biased towards one label. These findings emphasize the importance of further systematically investigating numerical reasoning capabilities across various NLP models. Our \textbf{contributions} are as follows:


    \begin{itemize}
        \item We introduce a taxonomy for numerical reasoning skills, including representation/number sense/manipulation skills and complex reasoning with numbers. 
        \item We propose a semi-automated approach to create large-scale, numerical probe sets.
        \item We evaluate three different categories of language models (LMs) on our numerical probe sets: $(i)$ numerical LMs; 
        $(ii)$ LMs for tabular data; 
        and $(iii)$ zero-/few-shot LLMs. 
    \end{itemize}


\section{A Taxonomy for Numerical Reasoning}
\label{sec:taxonomy}


This section introduces a hierarchical taxonomy for numerical reasoning, inspired by previous works on numeracy in NLP ~\citep{thawani-etal-2021-representing,xu-etal-2022-towards-robust} and cognitive science ~\citep{Barrouillet1998,Whyte2008,Bofferding2019}.
We group numerical reasoning skills given their complexity level in four categories:
$R1 - R4$.


\subsection{Number Representation ($R1$)}

\begin{figure}
\centering
\includegraphics[scale=0.62]{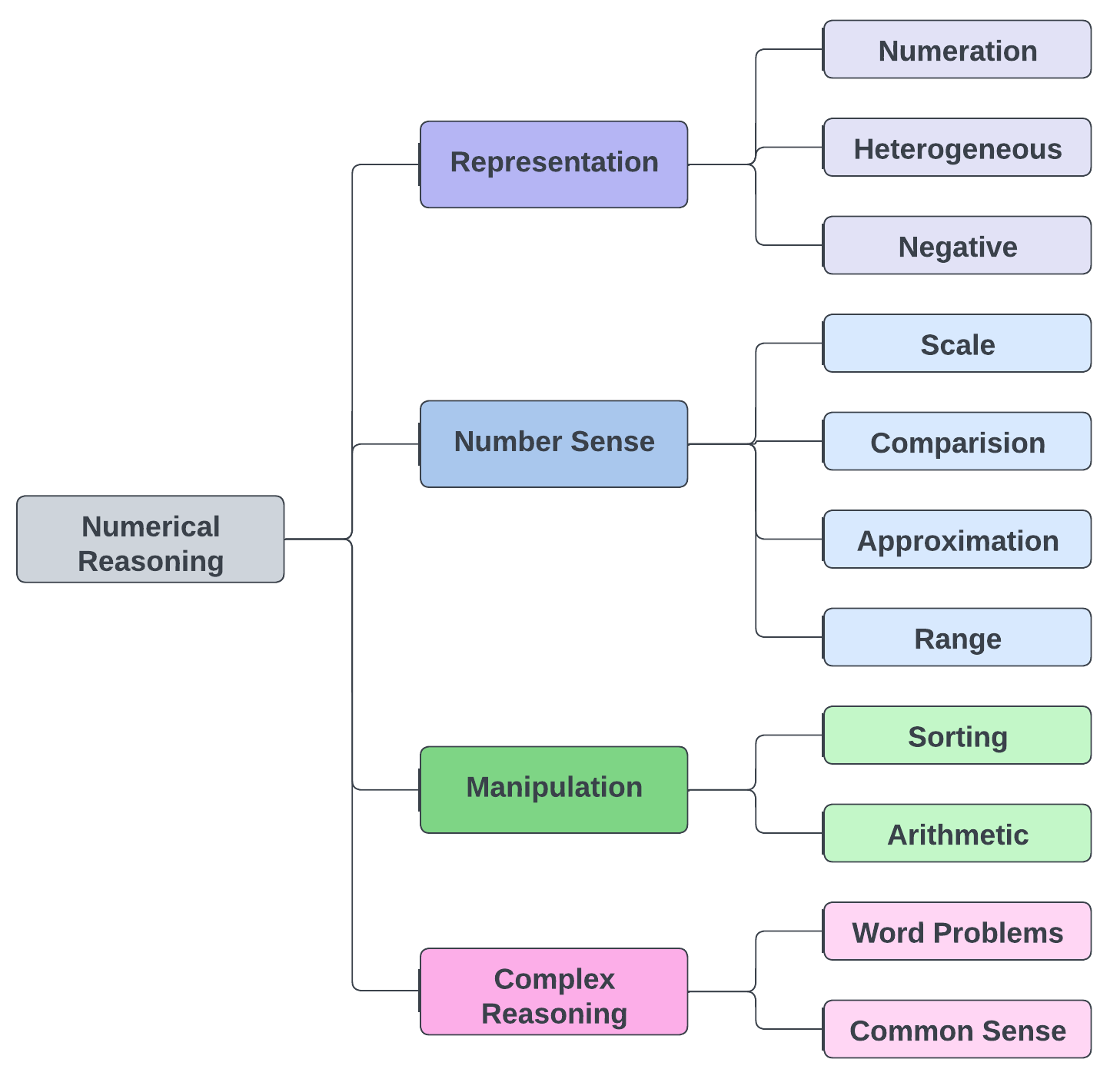}
\vspace{-1.0em}
\caption{\small  \label{fig:numreasoning} Overview of numerical reasoning types.}
\vspace{-1.5em}
\end{figure}

This category includes skills for understanding the \emph{form} of numerical data.
Similar to the notion of form in language~\citep{bender-koller-2020-climbing}, this is the realization of numbers in text; the way they are represented and expressed.

\paragraph{Numeration.} Numeration studies language model’s understanding of 
representation systems common for numbers in English:
numerical digits ($2$) and text (two). Specifically, we probe if LMs can link between distinct symbols used for the same number.
For example in Figure~\ref{fig:numreasoning}, H$3$ contains ``two'' as a word, which can be also represented through ``$2$''.

\paragraph{Heterogeneous Number Types.}
Formatted numbers (e.g. dates, times, and fractions) are frequently used to convey additional information associated with a numerical value. 
Numbers are formatted in a specific way given their context and purpose, 
such as expressing times and dates using full-stop (``.''), using the ``\%'' symbol to indicate fractions, and different currency symbols for money (i.e. ``\$'' or ``€''), e.g. H$1$ and ``Arith'' in Figure~\ref{fig:numreasoning}. 



\paragraph{Negative Numbers.}
Early on in their development, children develop some mental model for negative numbers (see experiments with first-graders in \citet{Bofferding2019}). Using negative numbers requires understanding the notation of negatives (i.e. ``$-$'' followed by a number). 
This also includes distinguishing between minus in subtractions ($1-3$), dates ($12$-$12$-$2022$), counts (i.e. from one to three) and in negative numbers ($-2$). 

\subsection{Number Sense ($R2$)}

Number sense includes reasoning skills for conceptualizing number quantities and understanding their meaning in a given context. 



\paragraph{Scale.}
In everyday communication, numbers commonly occur with measurement scales, e.g. weights, distances, or heights.
Understanding numbers in context of scales is a basis for various applications, e.g. question answering (e.g. \textit{``We are driving $80$ km$/$h, is this within the speed limit?''}), commonsense (e.g. \textit{``Cats weight between four and five kilograms.''})~\citep{lin-etal-2020-birds}, and temporal reasoning (e.g. \textit{``She left the office thirty minutes ago.''})~\citep{zhou-etal-2020-temporal, zhang-etal-2020-language-embeddings}. 

\paragraph{Comparison.}
Comparing numbers allows understanding numerical relationships. It involves identifying which numbers are greater than, less than, or equal to others. 
For example, given the table in Figure~\ref{fig:numreasoning}, 
understanding \emph{``The running time of Hulk is longer than $120$ minutes.''} requires comparison.  

\paragraph{Range.}
The question \textit{``Was the budget of the movie between $\$130$ and $\$245.4$?''} about the table in Figure~\ref{fig:numreasoning} requires understanding number ranges. 
Already at an age between two and three years, children develop numerical abilities to understand sequences of numbers and start reciting numbers in an appropriate order ~\citep{fuson2012children,Laski2007}. 
Models' that understand the notation of ranges, can correctly answer the question by knowing that $137$ is in the range $130 - 245.4$.
 


\paragraph{Approximation.}
Humans commonly approximate number values in everyday life ~\citep{Odic2018,Bonny2013}. 
H$3$ in Figure~\ref{fig:numreasoning} requires approximation among other skills to map ``about two hours'' to ``$138$ minutes'' in the table. 
As a reasoning skill, it allows to make quick estimations and metric unit conversations, and understand the approximate values of numbers without calculating them explicitly.


\subsection{Manipulation ($R3$)}

Manipulation reasoning types are used to apply basic operations on numbers such as addition. Successful manipulation of numbers requires understanding their \emph{representations} and meaning in the given context (i.e. \emph{number sense}). 

\paragraph{Sorting.}
The sentence \textit{``Out of all Ang Lee's directed movies, `Hulk' was the one with the second highest box office income.''} requires sorting all movies according to their box office income in order to select the second highest income. 
Sorting objects given criteria is a basic milestone for developing cognitive skills. By age two, children already begin to understand the concept of sorting.

\paragraph{Simple arithmetic.}
Arithmetic reasoning is the ability of manipulating numbers with basic operations (addition, subtraction, multiplication, division). While adults commonly retrieve results of simple calculations from memory, children apply different operations~\citep{barrouillet1998algorithmic}. 


\subsection{Complex Reasoning ($R4$)}

This category builds on all previous reasoning categories ($R1 - R3$) to solve numerical word problems (NWP). 
NWP are expressed through natural language and require multistep reasoning. Extracting information from the problem description and applying numerical/mathematical reasoning using the retrieved information and world/commonsense knowledge is required~\citep{upadhyay-chang-2017-annotating, amini-etal-2019-mathqa, huang-etal-2016-well}.


\section{Numerical Probing Framework}
\label{sec:framework}

This section provides an overview of the probing framework. 
We use TNLI datasets for automated probe creation. 

\subsection{Preliminaries}

\paragraph{Tables for numerical probing.}

Tables align well with our objectives given three key criteria: $(i)$ 
numerical data is common in tables; $(ii)$ tables are frequent in real-world data sources; $(iii)$ tables, due to their structured formats, facilitate automated perturbations for probe creation. 
Tables' semi-structured format, the alignments available between table cells and column/row headers, and the frequency of numbers, make them well suitable for creating numerical probes automatically.

\paragraph{Table NLI.}
Given a natural language sentence as hypothesis and a tabular premise, the aim of TNLI is to classify if the hypothesis \emph{entails} or \emph{contradicts} the table~\citep{gupta-etal-2020-infotabs}. 
We use the table NLI datasets \emph{TabFact}~\citep{DBLP:conf/iclr/ChenWCZWLZW20} and InfoTabs~\citep{gupta-etal-2020-infotabs}, as well as recast the table QA datasets TAT-QA~\citep{zhu-etal-2021-tat} and TabMWP~\citep{lu2023dynamic} to NLI (i.e. \emph{TATQA-NLI}, \emph{TabMWP-NLI}). 
TAT-QA includes metadata, i.e. annotations of cells and operations per correct answer. This information is not available for any TNLI dataset and is crucial to create probes for specific reasoning types, e.g. \emph{arithmetic reasoning}. 
Table~\ref{tab:dataset_statistics} provides an overview of the TNLI datasets. 

\paragraph{Preprocessing.}
For each of numerical reasoning type, we first identify base TNLI hypotheses and/or tables in the datasets that can be used for automated probe creation. 
Hereby, we defined a list of reference tokens specific for each reasoning type and to filter relevant dataset samples. For example, we use units of measurements such as ``hour'', ``meter'', or ``kilogram'' filter hypotheses for \emph{scale} probes (see \S\ref{sec:probes} for more details).
To recast the TAT-QA dataset, we follow the simple yet effective, rule-based approach proposed by \citet{DemszkyGL18} for QA to NLI conversion.

\subsection{Probes through Structural Perturbation}
\label{ssec:hypothesis_probes}

Our framework includes three types of probes, created through hypotheses perturbation and counterfactual tables. 


\paragraph{1. Hypothesis label-preserving probes}
We create label-preserving probes changing the base hypothesis such that its meaning is not changed and the initial label is preserved. They are used to evaluate model's ability to reason and predict the correct label given semantically-equivalent changes.


\paragraph{2. Hypothesis label-flipping probes}

To generate label-flipping probes,
we modify the base hypothesis such that its meaning alters and the probe label flips, e.g. from entailment to contradiction. 
We aim to overcome potential dataset artefacts that might be exploited for label prediction instead of performing numerical reasoning.
These changes are specific to the reasoning type. For example, to flip labels of \emph{scale} probes, we substitute measurement units for a particular scale (e.g. ``kilograms'') by another unit (e.g. ``meters'') or introduce errors in conversion of units (e.g. $3,000$ meters instead of $3$ kilometers).  
%


\begin{table}
\centering
\scalebox{0.76}{
\begin{tabular}{l l l l l}
\hline
\textbf{Dataset}  & \textbf{Hypotheses} & \textbf{Tables} & \textbf{Num cells} & \textbf{Probes}\\
\hline
TabFact  & 118,275 & 16,573  &  59.00\% & 214,440 \\
InfoTabs & 23,738 & 2,540 &53.6\% & 19,779 \\
TATQA-NLI & 4,947 & 2,156 & 59.7\% &15,139 \\
ToTTo & 1,000 & 892 & 45.7\%& 1,000 \\
TabMWP & 283 & 283 & 38.3\%  &238 \\
\hline
\end{tabular}}
\vspace{-0.5em}
\caption{\small \label{tab:dataset_statistics} TNLI probing datasets; \textit{num cells} refers to the average ratio of numerical cells in tables. 
}
\vspace{-1.75em}
\end{table}

\paragraph{3. Table Probes through Counterfactual Table Editing}

We also probe with counterfactual tables to evaluate if models rely on spurious patterns in the premise table for label prediction. 
We filter the counterfactual datasets by \citet{jena-etal-2022-leveraging} consisting of \emph{\{hypothesis; original table; counterfactual table\}} for numerical hypotheses. 


\section{Probing with TNLI Datasets}
\label{sec:probes}


\begin{table*}[t]
\small

\begin{center}
\parbox{.90\linewidth}{
\centering
\begin{tabular}{l | l}
\hline
\multicolumn{2}{c}{\bf Rafael Nadal} \\ 
\hline
\textbf{Plays} & Left-handed \\
\textbf{Born} & 3 June 1986 (age 37) \\
\textbf{Height} & 1.85 m \\
\textbf{Turned pro} & 2001 \\
\textbf{Prize money} & US\$116,111,561 (3rd all-time leader in earnings)\\
\hline
\end{tabular}}


\end{center}
\begin{center}
\vspace{-1.0em}
\parbox{0.9\linewidth}{
\centering
\begin{tabular}{|lll|}
\multicolumn{2}{c}{} \\
\hline

\textit{Base Hypothesis H\textsubscript{1}}& Born in \colorbox{green!25}{1986}, Nadal is age \colorbox{green!25}{37} currently. & \\
\textit{Numeration Probe H\textsubscript{1}}& Born in \textcolor{blue}{nineteen eighty six}, Nadal is age \textcolor{blue}{thirty seven} currently. &\\
\textit{Num Flip Probe H\textsubscript{1}}& Born in \textcolor{red}{nineteen ninety two}, Nadal is age \textcolor{red}{forty one} currently. &\\
\textit{Range Probe H\textsubscript{1}}& Born in 1986, Nadal is age \textcolor{blue}{between 31-43} currently. &\\

\multicolumn{3}{|c|}{} \\

\textit{Base Hypothesis H\textsubscript{2}}& The player's birth date is on  \colorbox{green!25}{3rd June, 1986}. & \\
\textit{Heterog Probe H\textsubscript{2}}& The player's birth date is on  \textcolor{blue}{03-06-1986}. & \\
\textit{Heterog Flip Probe H\textsubscript{2}}& The player's birth date is on  \textcolor{red}{15-01-1999}. & \\
\multicolumn{3}{|c|}{} \\

\textit{Base Hypothesis H\textsubscript{3}}& With \colorbox{green!25}{\$116,111,561} prize money, he is the \colorbox{green!25}{3rd} highest earning all-time player. & \\
\textit{Heterog Probe H\textsubscript{3}}& With \textcolor{blue}{\$$116.111561e-6$} prize money, he is the \textcolor{blue}{third} highest earning all-time player. & \\
\textit{Approx Probe H\textsubscript{3}}& With \textcolor{blue}{about \$$116,000,000$} prize money, he is the 3rd highest earning all-time player. & \\

\multicolumn{3}{|c|}{} \\

\textit{Base Hypothesis H\textsubscript{4}} & Rafael Nadal has a height of \colorbox{green!25}{1.85 meters.} & \\
\textit{Scale Probe H\textsubscript{4}} & Rafael Nadal has a height of \textcolor{blue}{185 centimeters}. & \\
\textit{Scale Flip Probe H\textsubscript{4}} & Rafael Nadal has a height of \textcolor{red}{5.2 ft}. & \\
\multicolumn{3}{|c|}{} \\

\textit{Base Hypothesis H\textsubscript{5}} & \colorbox{green!25}{After} the year \colorbox{green!25}{2000,} the player Nadal turned pro.  & \\
\textit{Comparison Probe H\textsubscript{5}} & \textcolor{blue}{After} the year \textcolor{blue}{1995}, the player Nadal turned pro.  & \\
\textit{Comparison Flip Probe H\textsubscript{5}} &  \textcolor{red}{Before} the year \textcolor{red}{1990}, the player Nadal turned pro. & \\
\multicolumn{3}{|c|}{} \\

\hline
\end{tabular}
}
\end{center}
\vspace{-1.0em}
\caption{\small Exemplary hypotheses and non-/flipping probes for evaluated reasoning types}
\label{tab:probe_examples}
\vspace{-1.5em}
\end{table*}



This section discusses probe creation for each reasoning type from \S\ref{sec:framework} in detail.\footnote{Find details on probe statistics in Appendix~\ref{sec:probe_statistics}.}

\paragraph{Numeration.}
To study models' understanding of string (two) and numerical (e.g. $2$) number representations, we create two types of numeration probes. 
Once, filtering hypotheses with numbers written as strings (two) and substitute them by their numeric counterparts, while the second category applies the conversion vice versa.
The label-preserving probes are semantically equivalent to the base hypotheses and the label (e.g. \textit{entailment}) is not changed. 
Label-flipping probes replace the converted number $x$ by a random number in the range of $[x-x*0.5; x+x*0.5]$.
For example, the numeration flipping probe of H$1$ (Table~\ref{tab:probe_examples}) replaces \colorbox{green!25}{$112$} by \textcolor{red}{one hundred and forty-four} and flips the label from \textit{entailment} to \textit{contradiction}.

\paragraph{Heterogeneous number types.}
We create heterogeneous probes for the following categories frequent in the TNLI datasets: date formats, ordinals, percentage, currencies, and scientific notation.
To filter base hypotheses, we apply rule-based approaches specific to each category (i.e. dates formats, percentage, ordinals, etc.). 
To create label-preserving probes we apply representation-level changes which do not change their semantic meaning. For H$3$, we substitute \colorbox{green!25}{3rd June, 1986} by another English date format \textcolor{blue}{03-06-1986}. To flip the label, we replace the date in the adjusted format by a random date, i.e. \textcolor{red}{15-01-1999}.
We replace percentage signs by the token \textit{``percentages''} and vice versa. Similarly, ordinals written as words (\textit{first}) are exchanged by numerical representations ($1st$) and the other way around. For hypotheses with large numbers (e.g. ``\colorbox{green!25}{\$116,111,561}'' in H$3$), we introduce scientific notations (\textcolor{blue}{\$$116.111561e-6$}).

\paragraph{Negative numbers.}
To create negative probes, we replace negative numbers $-n$ (e.g. $-3$) by string equivalents 
(e.g. \emph{minus $3$}; \emph{negative $3$}).
For label-flipping probes, we convert negative numbers into the positive counterpart $n$. For example, converting \textit{``The company's monthly closing resulted in \colorbox{green!25}{-5} million USD.''} to \textit{``The company's monthly closing resulted in \textcolor{red}{5} million USD.''} flips the label. 

\paragraph{Scale.}
We create two types of scale probes: \emph{$(i)$ conversion} and \emph{$(ii)$ mapping}. 
Conversion converts numbers within a measurement scale. 
For H$4$ in Table~\ref{tab:probe_examples}, we convert the number and measurement unit (\colorbox{green!25}{1.85 meters}) to the next smaller unit within the same scale (\textcolor{blue}{185 centimeters}) for the label-preserving probe. 
For label-flip, we introduce an error in the converted number, i.e. converting \colorbox{green!25}{1.85 meters.} to \textcolor{red}{5.2 ft} instead of \textcolor{blue}{6.07 ft}.
Mapping probes replace the number and measurement unit by an equivalent (e.g. \colorbox{green!25}{$1.85m$} by \textcolor{blue}{$1.85$ meters}) for label-preserving probes and a random measurement unit e.g. \colorbox{green!25}{$1.85m$} to \textcolor{red}{$1.85$} kilograms) to flip the base hypotheses. 

    
\paragraph{Comparison.}
We first created a list of signal word-pairs by prompting GPT$3.5$. The list includes pairs such as \{``bigger'':``smaller''\}, \{``taller'':``shorter''\}, and \{``faster'':``slower''\}. 
Using these pairs and their synonyms, we filter base hypotheses and create three types of comparison probes. 
First, changing the signal word with its opposite counterpart to flip labels (see H$5$ in Table~\ref{tab:probe_examples} flipping \colorbox{green!25}{``after''} to ``\textcolor{red}{before}''). 
Second, altering the number such that the comparison and label do not change:
replacing \colorbox{green!25}{``after 2000''} by ``\textcolor{blue}{after 1995}'' ($H5$). 
Finally, we combine both prior approaches to create label-flipping probes, e.g. ``\textcolor{red}{Before} the year \textcolor{red}{1990}, the player Nadal turned pro.''
    
\paragraph{Approximation.}
Given the value of a number $n$ occurring in the base hypothesis, we decide the magnitude of rounding to apply for approximation probes. Smaller numbers are rounded to tens and larger ones to hundreds, thousands or larger decimal points.
For example, we create the probe \textit{``With \textcolor{blue}{about \$$116,000,000$} prize money, he is the 3rd highest earning all-time player''} by rounding \colorbox{green!25}{\$116,111,561} to ``\textcolor{blue}{about \$$116,000,000$}'' (H$3$). 

\paragraph{Range.}
To create range probes, we substitute a number $n$ in the base hypothesis by an appropriate range, e.g. \colorbox{green!25}{37} by ``\textcolor{blue}{between 31-43}'' (H$1$). 
We define the radius of the range and its boundaries given the value of $n$. 
For example, for $n<10$, we randomly sample a radius between $1-5$. For $n=7$ and a sampled radius of $2$, the range will be $[5-9]$.
We select decimal boundaries if $n$ is a decimal number. 

\paragraph{Sorting.}
We utilize table columns as number sequences to create sorting probes.
We generate a list of position indicators in number sequences (e.g. ``top'', ``second'' ``3rd'',``biggest'', ``lowest''). 
These words are used to filter base hypotheses. 
To create label-flipping probes, we change the position of the sequence to another one. For instance, we modify ``in the \colorbox{green!25}{first} quarter of 2018'' to ``in the \textcolor{blue}{third} quarter of 2018'' by selecting the value from the third row instead of the first.


\paragraph{Simple arithmetic.}
Using metadata on numbers and numeric operations for each hypothesis, we create arithmetic probes using the TATQA-NLI dataset.
We extract probes involving addition, subtraction, multiplication, and division. Additionally, we generate label-flipping probes by replacing the operation output (e.g. result of subtraction) in the hypothesis with a different number.
In Table~\ref{table:intro}, the \emph{``Arith''} probe involves calculating the difference between the \textit{budget} and \textit{box office} values to determine the correctness of \textcolor{blue}{$108.4$}. The flipped arithmetic probe produces a close but incorrect subtraction output, \textcolor{red}{$120.9$}.
    
\paragraph{Numerical word problems.}
We convert TabMWP questions-answers into declarative hypotheses. TabMWP is a math word problem dataset that involves reasoning with tables. For label-flipping probes, we replace numbers in the hypotheses with other numbers from the same column.


\paragraph{Counterfactual Table NLI Probes.}

We filter the counterfactual ToTTo~\citep{parikh-etal-2020-totto} dataset by \citet{jena-etal-2022-leveraging} for numerical hypothesis. 
To create counterfactual tables, they swap two or more table cells to modify the tables such that the label of the respective hypothesis changes from entailment to contradiction and vice versa. 


\section{Experiments and Analysis}

Next, we provide an overview of all probed models. We also discuss the obtained results and insights.

\begin{table*}
        \centering
        \scalebox{0.7}{
        \begin{tabular}{l r r r | r r r r | r r r r r r}
        \hline
        \bf Model	& \multicolumn{3}{c|}{\bf Table Specific} & \multicolumn{4}{c|}{\bf Numerical Specific} & \multicolumn{6}{c}{\bf Large LMs}\\
        \hline
         \bf  & \bf  TAPAS & \bf  	DeBERTa &  \bf TAPEX &  \bf 	NT5 & \bf 	LUNA & \bf 	PASTA & \bf  	ReasTAP &	\multicolumn{2}{c}{\bf FlanT$5$}   & \multicolumn{2}{c}{\bf GPT$3.5$} & \multicolumn{2}{c}{\bf PaLM}  \\
          \bf Reasoning & \bf & \bf &  \bf &  \bf & \bf & \bf & \bf & \bf	few &	 \bf  zero & \bf 	 few &	 \bf  zero & \bf  few &	 \bf zero \\
        \hline
        \multicolumn{14}{c}{\bf Representation} \\
        \hline
        Numeration &-0.32 &-1.82 &-7.84 &-4.18 &-5.22 &-7.7 &-7.18 &1.28 &-8.84 &-0.47 &5.65 &-1.35 &-3.07 \\
        Heterogeneous &-4.03 &-2.36 &-5.94 &-3 &-10.09 &-7.76 &-3.18 &0.34 &-5.49 &6.8 &6.65 &0.44 &-2.22 \\
        Negative  &-46.11 &-13.77 &0.56 &-94.48 &-75.55 &-10.68 &2.65 &19.21 &42.3 &8.24 &2.3 &-2.17 &1.14 \\
        \multicolumn{14}{c}{\bf Label Flipped} \\
        Numeration &-38.87 &4.09 &-43.3 &-48.53 &-71.35 &-25.85 &-37.21 &-78.37 &33.38 &-37.78 &44.71 &-37.29 &-46.45 \\
        Heterogeneous  &-9.57 &8.53 &-32.25 &-1.97 &-43.48 &-23.59 &-16.21 &-53.44 &86.6 &-27.97 &27.79 &-20.65 &-25.31 \\
        Negative &-64.81 &-41.56 &-97.01 &-17.87 &76.85 &-70.58 &-96.46 &-83.92 &173.14 &-63.64 &2.2 &-80.43 &-78.41 \\
        \hline
        \multicolumn{14}{c}{\bf Number Sense} \\
        \hline
        Scale &0.03 &-6.25 &-12.91 &1.21 &-11.43 &-1.56 &-4.6 &-9.45 &-7.05 &2.46 &-0.52 &-3.71 &-17.58 \\
        Comparison &-21.8 &-18.18 &-12.58 &-29.19 &-30 &-35.11 &-40.02 &29.38 &140.82 &-9.39 &9.13 &-16.91 &-20.08 \\
        Approximation &-5.61 &-6.65 &-18.9 &-9.55 &-7.67 &-27.44 &-7.81 &-9.66 &-12.94 &0.03 &12.08 &-10.44 &-12.03 \\
        Range &-18.89 &-33.77 &-1.96 &-20.43 &-86.77 &-84.66 &4.97 &22.44 &178.13 &0.5 &-1.07 &14.37 &4.41 \\
        \multicolumn{14}{c}{\bf Label Flipped}\\
        Scale &-23.73 &-64.58 &-30.41 &-39.08 &-68.44 &-51.66 &-16.54 &-69.56 &93.77 &-39.08 &39.98 &-17.85 &-27.4 \\
        Comparison &57.67 &-19.36 &-4.83 &-29.62 &-0.28 &-19.1 &-15.65 &-8.47 &-40.75 &-20.81 &14.67 &-17.96 &-16.09 \\
        \hline
        \multicolumn{14}{c}{\bf Manipulation} \\
        \hline
        Sorting &-34.8 &28.66 &-91 &-22.6 &54.31 &-4.9 &-83.96 &-86.67 &25 &-57.39 &-5.62 &-32.45 &-39.59 \\
        Arithmetic  &-58.62 &-24.96 &-95.53 &-27.1 &7.07 &-49.06 &-88.87 &-71.53 &265.07 &-60.34 &4.87 &-67.67 &-64.98 \\
        \hline
        \multicolumn{14}{c}{\bf Complex Reasoning} \\
        \hline
        Complex &63.37 &6.93 &-80.18 &-3.22 &41.41 &-50.84 &116.17 &-89.77 &-40 &-60.22 &-4.35 &-73.4 &-75.9 \\
        Counterfactual &44.5 &55.54 &-12.29 &159.3 &0.98 &-6.09 &-10.12 &61.5 &12.23 &40.63 &5.26 &30.57 &48.56 \\
        \hline
        \end{tabular}}
        \vspace{-0.5em}
        \caption{\small \label{tab:probe_results} Probing results given as accuracy difference (in \%) between base hypotheses and probes.}
        \vspace{-1.5em}
        \end{table*}



\subsection{Probed Models}


We use state-of-the-art models which are divers in terms of architecture, size, and training setup, grouped into three categories:

\paragraph{$(C1)$ Numerical LMs.}

This category includes LMs adapted for numerical reasoning.
\emph{LUNA} ~\citep{HanXZSHZ2022} is a recent transformer-based model with an adapted tokenization approach for numbers. The model encodes numbers as single tokens (e.g. $3,201$) instead of splitting them down into subwords or binned tokens. 
\emph{NT$5$} ~\citep{YangCCC2021} is a variation of the T5 model. It has been modified for numerical reasoning through additional pretraining objectives and fine-tuning using numerical datasets. 
\emph{PASTA} ~\citep{gu-etal-2022-pasta} is based on DeBERTa and is pretrained with objectives that use table-based numeric operations. 
ReasTAP~\citep{zhao-etal-2022-reastap} is a BART-based model pretrained on synthetically generated data requiring seven table reasoning skills, including a numerical task, temporal reasoning, and conjunction.

\paragraph{$(C2)$ LMs for tabular reasoning.}
\emph{TAPAS}~\citep{herzig-etal-2020-tapas} extends the BERT encoder with table-specific embeddings.
We used a TAPAS model trained with intermediate pretraining on synthetic and counterfactual data~\citep{eisenschlos-etal-2020-understanding}.
We also probe TAPEX~\citep{LiuCGZLCL22}, which uses BART~\citep{lewis-etal-2020-bart} as its base model and pretrains the model to mimic a neural SQL executor over tables.
Previous works have also shown the success of the \textit{*BERT} models on tabular NLI tasks~\citep{herzig-etal-2020-tapas,yin-etal-2020-tabert,shankarampeta-etal-2022-enhancing, akhtar-etal-2022-pubhealthtab}. Tables are either linearized or processed into sentences or structured formats. The transformed tables are then used as input to the models.
We use a DeBERTa model~\citep{HeLGC21} trained on multiple NLI datasets for this setting.

\paragraph{$(C3)$ Large LMs.}

For few-/zero-shot evaluation, we select FlanT$5$~\citep{Flan23}, GPT$3.5$, and PaLM $2$~\citep{PaLM23} and probe them in few-shot and zero-shot settings. 


\subsection{Training and Evaluation}
 
We fine-tune models 
with the base hypotheses from the training sets and evaluate models only on probes created with their test sets. 
The few-shot models are prompted with $2$-shot extrapolation. 
We evaluate all models in a $3$-step process: $(1)$ evaluation of base hypotheses $H$; $(2)$ evaluation of probes $P$, created using $H$; $(3)$ calculating changes in model performance by comparing accuracy of $P$ to $H$.
As our TNLI task is a binary classification task, we measure shifts in accuracy for evaluation. 

\subsection{Results and Discussion}


Table~\ref{tab:probe_results} gives on overview of all probing results. 
If available, we separately list scores for flipped probes, e.g. \emph{numeration} and \emph{numeration flipped}. 

\paragraph{$(Q1)$ Does any model excel in all numerical reasoning types?} 
While there is not one best-performing model across all reasoning types and different models struggle with different types, 
FlanT$5$ and GPT$3.5$ show overall good performance in a zero-shot setting. 
While GPT$3.5$ (fewshot) performance drops by $-60.22\%$ for complex reasoning probes, the model's average accuracy change is around $-16.7\%$ for other types. 
This can be related to $(1)$ models pretraining data, and $(2)$ training on chain-of-thought reasoning tasks \citep{wei2022chain}. 
GPT$3.5$ was trained on more than $300$ TB data Common Crawl, allowing the model to see much more (numerical) data during training than other probed models. In comparison, DeBERTa was trained on only $78GB$ of data~\citep{HeLGC21}.
Interestingly, both NT$5$ and FlanT$5$ use T$5$ as their base model. FlanT$5$ was instruction-fine-tuned and outperforms NT5 in many probing categories. 


\paragraph{$(Q2)$
What factors can contribute to high performance variations across certain reasoning types?}
Large performance variations mainly occur due to inconsistent numerical reasoning of models across tasks. For example, we observe that some models struggle with more basic reasoning (e.g., FlanT5 zero on numeration) while performing very well on more complex types.
This behavior might have different reasons. One potential reason is memorization. Previous works~\citep{petroni-etal-2019-language, CarliniTWJHLRBS21, Ishihara2023} show that large pretrained language models store knowledge in their parameters, which they tend to retrieve instead of reasoning over the provided input~\citep{gupta-etal-2022-model}. Hence, models can memorize common arithmetic operations they encounter during training and perform well on certain downstream tasks. For example, flipping numbers as words (``two'') to numerals (``$2$'') might allow models to retrieve knowledge which they didn't consider for the initial hypothesis. 
%
Another reason for high-performance drops can be the hallucination of models. While models  initially perform well on hypotheses, adjusting the numbers can hinder models from relying on spurious patterns.

\paragraph{$(Q3)$
How do models perform on different types of numerical reasoning?}

{\bf Representation.} 
In Table~\ref{tab:probe_results}, comparing numeration probes, we find for all models a performance drop of between $[0; -10]$ percentages for numeration probes, except FlanT$5$ (few).
This drop strongly increases for almost all models evaluated on numeration flipped probes. For example, FlanT$5$ (few) shows a performance drop of $-78.37\%$.
%
FlanT$5$ (few) also performs well on heterogeneous probes, followed by DeBERTa ($-2.4\%$) and NT5 ($-3\%$). 
Whereas LUNA performance drops significantly for heterogeneous probes (flipped and non-flipped).
TAPAS, NT5, and LUNA show significant performance drops (between $-38.87\%$ and $-71.35\%$) on negative number probes. This could be because the models exploit correlations between the ``$-$'' sign and labels for predicting base hypotheses. Interestingly, few- and zero-shot models like FlanT$5$ and GPT$3.5$ show improvements on negative number probes. This might be attributable to models understanding probed versions of negative numbers (e.g. ``minus $22$'') better than the representation in the initial hypotheses (e.g. ``$-22$'').



{\bf Number sense.} 
Comparing models based on number sense probes, we observe different patterns for fine-tuned models and few-/zero-shot models. Fine-tuned models struggle especially with comparison probes, with a $-26.7\%$ average performance drop. Scale probes show a $-42.1\%$ decrease on flipping probes, while approximation (flipping) probes report a $-12.0\%$ decrease in model performance. In contrast, FlanT$5$ perform better on comparison and range probes, sometimes surpassing predictions on the base hypotheses. All models demonstrate lower performance on approximation probes compared to the base hypotheses, with PASTA performance dropping by $-27.44\%$.

{\bf Manipulation and Complex Reasoning.}
Fine-tuned models exhibit an average accuracy drop of $-57\%$ on arithmetic probes, except for LUNA with a slight performance increase. The performance of PaLM (zero) and FlanT$5$ (few) drops by $-67.67\%$ and $-71.53\%$, respectively.
All models' performance drops on sorting probes (avg. $-27\%$), except for DeBERTa, LUNA, and FlanT$5$ (zero).
Unlike most other reasoning types, fine-tuned models outperform few-/zero-shot models on complex reasoning probes. ReasTAP achieves the highest accuracy, followed by TAPAS and LUNA. FlanT$5$, TAPEX, and PaLM have the largest performance drops on complex reasoning probes.

\paragraph{$(Q4)$ Do models perform similarly for flipped and non-flipped probes?} 
We observe higher performance drops for label-flipping probes compared to non-flipping probes across models. Models that struggle with flipping probes but perform well on their non-flipping counterparts indicate a reliance on spurious patterns for label prediction. 
The performance of TAPAS, TAPEX, PASTA, ReasTAP, FlanT$5$ (few), and PaLM drops significantly for the representation reasoning category comparing non-flipping and flipping probes.  
For example, TAPAS performance drops by $-2.28\%$ on numeration probes, but show a drop of $-45.98\%$ on numeration flipping probes. Similarly, DeBERTa performs well on scale probes ($-6.25\%$) compared to the flipping version ($-64.58\%$). PaLM performance on numeration, heterogeneous, and negative probes drops by approximately $-35\%$, $-20\%$, and $-80\%$ on flipping counterparts. 
DeBERTa exhibits robust performance on number flipping probes for sorting and FlanT$5$ on negative numbers, as well as arithmetic probes.

\paragraph{$(Q5)$ Are numerical and table-specific models better for numerical reasoning than LLMs?} 

{\bf Numerical models.} 
Our experiments do not indicate any superiority of numerical models over others.
LUNA, a transformer model that uses a specific tokenization method for numbers, performs similarly to other models on many reasoning types. The only reasoning type where LUNA outperforms is comparison flipping probes, with a small improvement of $0.28\%$.
PASTA is a DeBERTa-based model trained on numerical data and pretraining objectives. 
However, compared to DeBERTa, it only performs better on negative number and scale probes. 
ReasTAP outperforms other trained models on negative probes but seems to rely on spurious patterns for its predictions as the performance drops significantly ($-96.46\%$) for negative flipping probes.


{\bf Table-based models.}
DeBERTa outperforms all other trained models on numeration and heterogeneous probes. For flipping probes, we also find the model to perform robustly except for negative number flipping.
Compared to other models, TAPAS performs well on heterogeneous probes, non-flipping scale probes, and complex reasoning probes. 
Table-based models do not perform better than other models on number sense and manipulation.

{\bf LLMs.} 
Comparing to other model categories, LLMs perform well on non-flipped representations and number sense probes but show large performance drops for most flipped counterparts. Except FlanT$5$ (zero), models also show a drop of more than $-30\%$ on sorting. Different to other model cateories, the accuracy drops across all LLMs for complex probes (by $-67.7\%$ in average).

\section{Related Work}
\label{sec:relwork}

\paragraph{Numeracy Taxonomies in NLP.}

\citet{thawani-etal-2021-representing} categorise NLP work on numeracy along the dimensions of granularity (i.e. exact and approximate numbers) and unit (i.e. abstract and grounded numbers). 
\citet{xu-etal-2022-towards-robust} evaluate the robustness of QA systems in handling numerical data with two probing tasks: $(i)$ numerical parsing and $(ii)$ semantic parsing.
The DROP QA benchmark~\citep{dua-etal-2019-drop} requires discrete operations (e.g. subtraction, count, sort) 
to answer questions over text.
While \citet{thawani-etal-2021-representing} concentrate on number representations in NLP systems, our work includes three further numerical reasoning categories. 
\citet{xu-etal-2022-towards-robust} focus on the robustness of NLP models in handling numerical data.
Our probing study on the other side pivots towards the reasoning capabilities of models when dealing with numerical and tabular data.
Different to prior work, our study gives a broad and detailed evaluation of ten models
with diverse architectures, sizes, and training settings for numerical reasoning on more than ten different reasoning types, ranging from representation to complex reasoning.







\paragraph{Language Model / Numerical Skills.}

Various studies have evaluated LMs' numerical skills in recent years. 
Earlier works probed word embeddings for numeration (e.g. $4$=four)~\citep{naik-etal-2019-exploring}, comparison (e.g. $3<4$)~\citep{wallace-etal-2019-nlp}, scale~\citep{zhang-etal-2020-language-embeddings}, and superlatives~\citep{wallace-etal-2019-nlp}. 
More recent works evaluate LMs on out-of-distribution numbers \citep{kim-etal-2021-seen}, numeration/magnitude/sorting/superlatives~\citep{pal-baral-2021-investigating-numeracy}, and arithmetic~\citep{MuffoCB22}. 
Our work builds upon these previous evaluation studies and extends them with further numerical probing categories, e.g. heterogeneous numbers.


\paragraph{Numerically-tuned Language Models.}
Various numerical LMs have been developed in recent times. 
\citet{geva-etal-2020-injecting} and \citet{liang-etal-2022-mwp} inject numerical skills into BERT through numerical pretraining objectives. 
PASTA~\citep{gu-etal-2022-pasta} and NT$5$~\citep{YangCCC2021}, which are based on DeBERTa and T5 respectively, fall into the same category of models. 
Another line of work adjusts LMs' architectures for numerical reasoning through numerical tokenization~\citep{HanXZSHZ2022} or additional, numerical embeddings~\citep{JinJWLWRQ2021}. 

\paragraph{Systematic Probes for Tables.}

Tables have been previously used to create probes for table grounding~\citep{gupta-etal-2022-right} or recasting non-NLI datasets (e.g. question-answering) to NLI~\citep{jena-etal-2022-leveraging}. 
Unlike unstructured text data, tables have a natural structure that allows creating controlled experiments more easily~\citep{gupta-etal-2022-model}. We drew inspiration from prior tabular probing approaches and extended them for semi-automating creation of numerical probes. 
\citet{jena-etal-2022-leveraging} introduce a generic approach to recast table QA datasets to NLI data. 
For data recasting, they follow a systemic approach similar to ours. However, their focus is on transforming QA datasets, emphasizing the end-result (i.e. the NLI data) through data augmentation. 








\section{Conclusion}

This paper presents a framework for probing language models' numerical reasoning skills. We organise skills in a taxonomy and generate large-scale sets of probes covering more than ten numerical reasoning types.
Using table NLI as a case study, we evaluate the numerical reasoning abilities of ten models. These models belong to the categories numerical LMs, tabular LMs, and few-/zero-shot LLMs.
We discuss reasoning types that prove challenging for the probed models and explore promising directions for future research.

\textbf{Future Directions.}
For certain (numerical) tasks, tool-augmented LMs equipped with capabilities such as calculators or code execution have been proven valuable~\citep{MialonDLNPRRSDCGLS23}.
However, many tasks require implicit numerical reasoning which might not necessarily involve direct calculations based on numbers. For instance, classifying sentences that incorporate numbers in varied settings, like time indications~\citep{feng-etal-2023-generic}, currencies or conversations~\citep{Macinaetal23}. Such tasks demand a numerical interpretation beyond mere arithmetic computations.
Moreover, calling external tools using LMs requires basic numerical comprehension to invoke an external tool correctly~\citep{ChenMWC23}.


\section*{Limitations}

This work proposes a taxonomy and framework to probe numerical reasoning skills in LMs. It involves the creation of large-scale probing sets using an automated approach. However, the evaluation of this approach is currently limited to the task of table NLI.
For future research, it is interesting to extend this to include additional tasks and datasets. This extension serves two purposes: first, it allows evaluating a more diverse range of datasets. Second, it enables including challenges specific to other tasks.

In this paper, the evaluation of most reasoning types primarily involves structural changes at the hypotheses level. While we include counterfactual table probes, they are limited to one dataset and perturbations method only. 
Further research is needed to study models' performance on numerical data in the premise data. Therefore, we need table-based probes for all reasoning types of the proposed taxonomy.

\section*{Ethics Statement}

In this paper, we study the numerical reasoning skills of different LMs. However, to deploy these systems in real-world applications, further studies and evaluations specific to the intended use cases are required.
In order to support future research, we plan to release all the scripts and resources used for probe creation and model evaluation. This will facilitate and encourage further research in this field.


\bibliography{anthology,custom}
\bibliographystyle{acl_natbib}

\appendix

\section{Insights}
\textbf{Main Insights. }We investigated the language models and found that LLMs like FlanT$5$ and GPT$3.5$ perform better than other models on various numerical reasoning tasks. When the labels are switched around and when dealing with negative values, we found that both table-based and numerical models had difficulty comprehending the data. In contrast, DeBERTa performs relatively well compared to models like LUNA and PASTA, which are tuned for improved numerical reasoning skills.

\section{Probe Statistics}
\label{sec:probe_statistics}

\begin{table}[!h]
\centering
\scalebox{0.83}{
\begin{tabular}{l r}
\hline
\textbf{Reasoning Type} & \textbf{Count}\\
\hline
Word Problems & 238 \\
Sorting & 379 \\
Counterfactual & 1,000 \\
Currency & 1,014\\
Negative & 3,316\\
Range & 4,208\\
Scientific notation & 6,274\\
Arithmetic & 8,082\\
Ordinal & 10,569\\
Percentage & 16,851\\
Date & 18,642\\
Approximation & 20,440\\
Comparison & 30,763\\
Numeration & 166,319\\
\hline
Total       &         288,095\\
Flipped probes      &         77,687\\
\hline
\end{tabular}}
\caption{\label{tab:probe_statistics} Breakdown of probes per reasoning type.}
\end{table}

Table~\ref{tab:dataset_statistics} gives an overview of probes per dataset. Most probes (i.e. $214,440$) are created from TabFact hypotheses as this is also the biggest dataset available, followed by InfoTabs ($19,779$). 
Table~\ref{tab:probe_statistics} provides a breakdown of probes per reasoning type.
In total, we have $286,857$ probes, of which $76,404$ are label-flipping probes.

In the ideal scenario with counterfactual tables, the models' performance should be similar to the performance on the original tables. However, we observed that TAPAS and DeBERTa's performance improved significantly, which leads to the conclusion that models are biased toward one label.

Overall no language model excels in all the numerical reasoning tasks. Surprisingly, models perform relatively well in complex tasks like Numerical Word Problems but struggle at simple reasoning tasks like numeration and comparison.

\begin{table*}
    \centering
    \scalebox{0.7}{
   \begin{tabular}{l r r r | r r r r | r r r r r r}
  \hline
  \bf Model	& \multicolumn{3}{c|}{\bf Table Specific} & \multicolumn{4}{c|}{\bf Numerical Specific} & \multicolumn{6}{c}{\bf Large LMs}\\
  \hline
   \bf  & \bf  TAPAS & \bf  	DeBERTa &  \bf TAPEX &  \bf 	NT5 & \bf 	LUNA & \bf 	PASTA & \bf  	ReasTAP & \multicolumn{2}{c}{\bf FlanT$5$}   &	\multicolumn{2}{c}{\bf GPT$3.5$} & \multicolumn{2}{c}{\bf PaLM}  \\
    \bf Reasoning & \bf & \bf &  \bf &  \bf & \bf & \bf & \bf & \bf	few &	 \bf  zero & \bf 	 few &	 \bf  zero & \bf  few &	 \bf zero \\
  \hline
  \multicolumn{14}{c}{\bf Representation} \\
  \hline
  Numeration & 19.02 & 61.69 & 74.83 & 64.19 & 63.57 & 85.43 & 75.35 & 64.59 & 67.58 & 77.6 & 61.83 & 76.38 & 70.76  \\ 
Heterogeneous & 79.25 & 56.61 & 81.17 & 66.85 & 70.71 & 89.11 & 76.49 & 70.17 & 71.04 & 78.82 & 61.26 & 76.75 & 69.85  \\ 
Negative & 59.38 & 69.81 & 97.01 & 51.77 & 24.66 & 70.44 & 81.04 & 70.8 & 20.37 & 85 & 87 & 92 & 88  \\ 
  \multicolumn{14}{c}{\bf Label Flipped} \\
  Numeration & 23 & 61.92 & 75.79 & 71.98 & 76.34 & 86.96 & 73.64 & 88.93 & 66.93 & 90.57 & 54.11 & 85.97 & 87.91  \\ 
Heterogeneous & 67.66 & 53.12 & 82.96 & 62.94 & 77.57 & 83.75 & 68.37 & 87.48 & 51.55 & 89.18 & 65.32 & 86.55 & 79.14  \\ 
Negative & 59.38 & 69.81 & 97.01 & 51.77 & 24.66 & 70.44 & 72.53 & 70.8 & 20.37 & 88 & 85 & 92 & 88  \\ 
  \hline
  \multicolumn{14}{c}{\bf Number Sense} \\
  \hline
  Scale & 74.11 & 67.29 & 78.32 & 63.05 & 72.24 & 77.82 & 31.1 & 61.25 & 69.59 & 79.17 & 69 & 78.04 & 71.15  \\ 
Comparison & 68.43 & 68.7 & 54.12 & 66.85 & 69.19 & 84.75 & 58.46 & 37.77 & 23.87 & 83.31 & 50.58 & 77.43 & 63.78  \\ 
Approximation & 75.49 & 64.65 & 71.64 & 63.3 & 67.09 & 86.08 & 58.66 & 62.28 & 78.09 & 75.13 & 59.62 & 76.88 & 75.35  \\ 
Range & 70.84 & 57.53 & 94.99 & 58.14 & 48.86 & 73.25 & 87.77 & 80.2 & 20.09 & 90 & 85.5 & 80 & 81.41  \\ 
  \multicolumn{14}{c}{\bf Label Flipped}\\
  Scale & 72.33 & 76.74 & 71.78 & 65.17 & 78.8 & 87 & 95.92 & 85.71 & 59.6 & 92.62 & 60.71 & 79.58 & 80.21  \\ 
Comparison & 53.55 & 70.75 & 38.29 & 68.64 & 69.76 & 85.74 & 42.69 & 86.12 & 49.66 & 96.05 & 37.34 & 82 & 54.86  \\
  \hline
  \multicolumn{14}{c}{\bf Manipulation} \\
  \hline
  Sorting & 68.71 & 49.98 & 94.65 & 69.83 & 51.11 & 69.87 & 83.33 & 75.44 & 61.4 & 91.44 & 77 & 68.94 & 70.5  \\ 
Arithmetic & 71.15 & 58.01 & 95.92 & 58.83 & 73.6 & 73.6 & 88.97 & 78.92 & 22.42 & 89.5 & 85.5 & 83.5 & 81.5  \\ 
  \hline
  \multicolumn{14}{c}{\bf Complex Reasoning} \\
  \hline
  Complex & 54.6 & 52.65 & 82.69 & 52.94 & 39.22 & 81.93 & 95.92 & 91.69	& 69.03 & 93 & 56.1 & 94 & 91.3  \\ 
Counterfactual & 59.71 & 43.23 & 88.7 & 80.98 & 81.3 & 83.52 & 86.9 & 35.46	& 62.5 & 58.18 & 69.09 & 55.38 & 53.85 \\ 
  \hline
  \end{tabular}}
    \vspace{-0.5em}
    \caption{\small \label{tab:probe_results_og} Results on original sets (average accuracy).}
    \vspace{-1.5em}
    \end{table*}

\begin{table*}
    \centering
    \scalebox{0.7}{
   \begin{tabular}{l r r r | r r r r | r r r r r r}
  \hline
  \bf Model	& \multicolumn{3}{c|}{\bf Table Specific} & \multicolumn{4}{c|}{\bf Numerical Specific} & \multicolumn{6}{c}{\bf Large LMs}\\
  \hline
   \bf  & \bf  TAPAS & \bf  	DeBERTa &  \bf TAPEX &  \bf 	NT5 & \bf 	LUNA & \bf 	PASTA & \bf  	ReasTAP & 	\multicolumn{2}{c}{\bf FlanT$5$}   & \multicolumn{2}{c}{\bf GPT$3.5$} & \multicolumn{2}{c}{\bf PaLM}  \\
    \bf Reasoning & \bf & \bf &  \bf &  \bf & \bf & \bf & \bf & \bf	few &	 \bf  zero & \bf 	 few &	 \bf  zero & \bf  few &	 \bf zero \\
  \hline
  \multicolumn{14}{c}{\bf Representation} \\
  \hline
  Numeration & 18.4 & 59.24 & 68.34 & 60.5 & 59.44 & 79.64 & 69.36 & 64.5 & 63.01 & 77.19 & 64.3 & 75.38 & 74  \\ 
Heterogeneous & 72.76 & 54.52 & 75.75 & 62.19 & 63.73 & 80.59 & 74.32 & 68.92 & 67.52 & 77.36 & 65.78 & 75.5 & 70.22  \\ 
Negative & 32 & 60.2 & 97.55 & 2.86 & 6.03 & 62.92 & 76.19 & 84.4 & 28.98 & 92 & 89 & 90 & 89  \\ 
  \multicolumn{14}{c}{\bf Label Flipped} \\
  Numeration & 10.13 & 43.68 & 39.41 & 33.89 & 29.82 & 53.3 & 41.05 & 17.34 & 70.04 & 56.83 & 73.37 & 52.54 & 45.91  \\ 
Heterogeneous & 63.1 & 42.45 & 49.44 & 57.08 & 47.97 & 68.75 & 56.58 & 40.57 & 80.34 & 62.16 & 78.83 & 70.12 & 62.18  \\ 
Negative & 20.9 & 40.8 & 2.9 & 42.52 & 43.61 & 20.72 & 41.28 & 11.38 & 55.63 & 32 & 86.87 & 18 & 19  \\  
  \hline
  \multicolumn{14}{c}{\bf Number Sense} \\
  \hline
  Scale & 71.68 & 49.43 & 68.38 & 60.39 & 62.21 & 73.63 & 67.23 & 53.12 & 62 & 80.46 & 66 & 75.15 & 59.39  \\ 
Comparison & 42.78 & 54.91 & 45.32 & 45.94 & 54.79 & 60.04 & 40.73 & 48.84 & 55.33 & 73.74 & 58.69 & 64.97 & 51.71  \\ 
Approximation & 62.94 & 58.65 & 56.01 & 56.75 & 61.56 & 61.55 & 51.26 & 56.23 & 67.65 & 74.96 & 66.1 & 69.25 & 66.4  \\ 
Range & 57.73 & 38.09 & 93.13 & 46.28 & 6.46 & 11.24 & 92.12 & 98.2 & 55.87 & 90.47 & 84.5 & 91.5 & 85  \\ 
  \multicolumn{14}{c}{\bf Label Flipped}\\
  Scale & 39.26 & 44.24 & 38.91 & 44.55 & 28.63 & 43.12 & 98.46 & 23.69 & 77.67 & 56.31 & 66.94 & 60.93 & 52.57  \\ 
Comparison & 62.59 & 57.28 & 37.15 & 46.9 & 66.75 & 69.72 & 34.74 & 66.43 & 50.29 & 75.22 & 54.6 & 69.43 & 44.98  \\ 
  \hline
  \multicolumn{14}{c}{\bf Manipulation} \\
  \hline
  Sorting & 44.8 & 63.75 & 8.47 & 53.89 & 77.79 & 64.54 & 13.29 & 10.03 & 76.75 & 38.75 & 72.81 & 46.38 & 42.31  \\ 
  Arithmetic & 71.15 & 43.52 & 4.28 & 42.89 & 37.5 & 37.5 & 9.88 & 22.45 & 81.08 & 35.5 & 89.5 & 27 & 28.5  \\  
  \hline
  \multicolumn{14}{c}{\bf Complex Reasoning} \\
  \hline
  Complex & 89.22 & 56.3 & 16.39 & 51.24 & 55.46 & 40.28 & 3.4 & 9.38 & 41.42 & 37 & 72.78 & 25 & 22  \\ 
  Counterfactual & 86.21 & 67.24 & 77.8 & 31.23 & 82.1 & 78.43 & 78.1 & 57.27 & 70.14 & 81.82 & 72.73 & 72.31 & 80 \\ 
  \hline
\end{tabular}}
\vspace{-0.5em}
\caption{\small Results on probed sets (average accuracy).} \label{tab:probe_results_probe}
\vspace{-1.5em}
\end{table*}

\end{document}